\documentclass[11pt]{article}

\usepackage[margin=1in]{geometry}
\usepackage{amsmath,amssymb}
\usepackage{booktabs}
\usepackage{graphicx}
\usepackage{microtype}
\usepackage{xcolor}
\usepackage[round]{natbib}
\usepackage[colorlinks=true,linkcolor=blue!60!black,citecolor=blue!60!black,urlcolor=blue!60!black]{hyperref}

\newcommand{\Xcal}{\mathcal{X}}

\title{Replicating TRACE:\\
A Practitioner's Guide to Its Threshold and Particle Budget}

\author{
  Alex Chadyuk\thanks{Corresponding author: \texttt{alex.chadyuk@lotusflare.com}} \quad
  Alicia Zhang \quad
  Roy Kucukates \\
  LotusFlare Inc.
}

\date{1 September 2026}

\begin{document}

\maketitle

\begin{abstract}
TRACE \citep{math2026trace} reads causal graphs over event types out of a
pretrained autoregressive sequence model by thresholding a per-position
conditional-mutual-information (CMI) estimate at a fixed $\tau$. We report
an independent replication of its headline
synthetic result: with $\tau$ selected on a validation split, mean
per-sequence $F1$ against exact interventional truth reaches $0.90$--$0.91$
at $|\Xcal| = 1000$ (the paper's Table~2: $\approx 0.91$) and
$0.86$--$0.91$ at every vocabulary size from 100 to 2000. Four findings
inform the practical application of TRACE. First, the
optimal CMI threshold is pinned to the truth margin rather than to any
constant: at every vocabulary size the errors at $\tau^{*}$ straddle the
$\delta = 0.05$ margin that defines ground truth---the true edges it
gives up sit just above $\delta$, the false edges it accepts, just 
below---and the blind optimum lands near $\delta/2$ times the estimator's
calibration, confirmed out of sample at $|\Xcal| = 5000$. Second, when
TRACE is operated at a single global threshold it mostly recovers a
direct, adjacent-influence graph: lag-1 true edges are
read at their generator KL and recalled at $0.97$--$0.99$, while true
edges at lag $\geq 2$ read orders of magnitude lower---the reading-scale price of
randomizing mediating positions, which an exact test of direct
causal effect requires when the truth is unknown. A per-lag threshold
family recovers a third to a half of lag-2 truth, and on lag-uniform
data TRACE recalls every lag at $0.40$--$0.87$ from one
validated threshold, trailing an atomic-intervention control by
$8$--$26$ pp at lags 3--6. Third, the default lag decay of the
synthetic benchmark the paper proposes in its appendix concentrates
$\approx 85\%$ of interventional truth at lag 1 and pushes the
remainder below the estimator's noise floor, so headline $F1$ there
certifies lag-1 recovery only and conflates the benchmark's skew with
the algorithm's own limit; a flatter decay puts recallable truth at
every lag and separates the two. Fourth, the
$F1$ observable saturates from $N = 2$ particles at the selected
threshold---a property of the threshold's margin over the noise floor,
not of the estimator, which converges as $N^{-1/2}$. We distill the
findings into concrete guidance for practitioners applying TRACE to new
data.
\end{abstract}

\section{Introduction}

TRACE \citep{math2026trace} belongs to a growing family of methods that read
causal structure directly out of a pretrained autoregressive model: a
decoder is trained on discrete event sequences, the CMI between a candidate
cause and each downstream position is estimated by particle-based
do-sampling, and edges are called by thresholding the resulting CMI at a
fixed $\tau$. Its central synthetic claim (paper's Table~2) is $F1 \approx 0.91$
against interventional ground truth at vocabulary size $|\Xcal| = 1000$ and
sequence length $L = 64$.

We replicated this claim from scratch and characterized how its two main
free parameters---the threshold and the particle count---behave across
regimes. The characterization surfaced two further findings. One is a
lag-graded recall limit of TRACE's staircase construction---its
whole-tail randomization of mediating positions, which an exact test
of direct causal effect requires. The other is a lag-1 skew of the
paper's synthetic benchmark that exaggerates this recall limit. 
Two properties of the study bear on how to read it. Ground truth
is exact: we reimplemented the paper's generator, so per-sequence
interventional truth is computed from its conditionals rather than
approximated. And every quantitative claim is made on at least two seeds
with the cross-seed gap published, under thresholds selected on
validation data and decision rules fixed before the runs (details in the
Reproducibility statement).

Our contributions are:

\begin{enumerate}
\item \textbf{Replication.} An independent reimplementation, scored the
  paper's own way against exact interventional truth, reproduces paper's Table~2 at
  the paper's regime (Section~\ref{sec:headline}) and at every vocabulary
  size in $\{100, 500, 1000, 2000\}$ (Section~\ref{sec:scaling}).
\item \textbf{The threshold is margin-pinned.} In our redundancy-matched
  worlds the blind validation optimum straddles the truth margin
  $\delta$: the errors at $\tau^{*}$ are margin-band pairs on both sides
  of $\delta$, the optimum decelerates toward $\delta/2$ times the
  estimator's calibration, and an out-of-sample probe at
  $|\Xcal| = 5000$ confirms the margin reading against a power-law
  extrapolation of our own sweep. The
  transferable object is therefore the validation-selection protocol the
  paper already licenses, and the authors concur (personal
  communication; see Acknowledgments) (Section~\ref{sec:scaling}).
\item \textbf{A lag-graded recall limit of the staircase construction.} At any
  single global threshold, on data whose causal strength decays with
  lag, TRACE tends to recover a direct, adjacent-influence graph: lag-1 true
  edges are read at their generator KL and recalled at
  $0.97$--$0.99$, while true edges at lag $\geq 2$ read decades
  (orders of magnitude) low---traced to the collapse of the observed
  effect token's factual
  probability when the staircase redraws every mediating position out
  of its factual context. The limit is graded, not absolute: a per-lag
  threshold family recovers a third to a half of lag-2 truth on the
  same runs, and an atomic-intervention instrument on lag-uniform data
  isolates the staircase's own share of the deficit as a moderate,
  lag-graded residual ($8$--$26$ pp at lags 3--6)
  (Section~\ref{sec:lag}).
\item \textbf{The default benchmark cannot show this limit.} The
  synthetic generator is specified in the paper's evaluation appendix
  and is not part of the algorithm; its hard-coded lag decay skews the
  benchmark toward lag 1 twice over---about $85\%$ of
  interventional-truth mass sits at lag 1, and the deep-lag remainder
  reads below the estimator's noise floor---so the headline $F1$ is
  attainable by lag-1 recovery alone. Validation against the default
  generator therefore certifies adjacent-influence recovery only, and
  conflates the benchmark's skew with the staircase's limit; a
  flat-decay variant (one generator parameter) separates them
  (Section~\ref{sec:benchmark}).
\item \textbf{Particle budget.} A particle sweep saturates
  from $N = 2$ at the selected threshold; we diagnose this as a
  threshold-margin property, not estimator saturation, and give a rule for
  choosing the particle budget (Section~\ref{sec:particles}).
\end{enumerate}

Section~\ref{sec:practitioners} collects the practical guidance.

\section{Related work}
\label{sec:related}

Selecting a sparsity-controlling free parameter by a data-driven procedure
rather than a fixed value is canonical---StARS \citep{stars-liu-2010}
made stability-based selection the default for graphical-model
regularization---and the sensitivity of headline benchmark numbers to
tuning and setup choices is well documented across the causal-discovery
literature \citep{varsortability-reisach-2021,sober-look-ng-2024,
benchmark-consistency-zhang-2026}. We claim no novelty for the lesson;
what we add is a measured account of how one named 2026 method's threshold
and particle budget behave across regimes, together with a replication of
its headline number. TRACE sits in a wider event-sequence
causal-discovery literature (e.g.\ CASCADE \citep{cascade-cuppers-2024},
THPs \citep{thps-cai-2024}, SHP \citep{shp-qiao-2023}, CAUSE
\citep{cause-zhang-2020}); an executed cross-method comparison is outside
this note's scope.

\section{TRACE and its evaluation protocol}
\label{sec:trace}

We summarize only what the replication needs; see \citet{math2026trace}
and the dissertation \citep{math2026thesis}.

\paragraph{Estimator.} A decoder-only autoregressive model is pretrained on
event sequences over a vocabulary $\Xcal$. For a sequence $s$ and a
candidate cause position $p$, TRACE estimates the CMI between the event at
$p$ and each position $q$ in a context window of size $c$ after $p$ by
comparing the model's predictive distributions under $N$ particle
do-interventions (the ``staircase'' construction): each particle redraws
$p$ \emph{and} every position between $p$ and $q$ uniformly, the mediating
draws shared across the two arms of the contrast. In intervention-calculus
terms the construction recovers the \emph{randomized controlled direct
effect} of $p$ on $q$ under the model's conditionals: each particle
contrast is a controlled direct effect at a randomized mediator setting
\citep[Def.~4.5.1, Eq.~(3.14)]{causality-pearl-2009}---the do-semantics of
graph surgery \citep[Def.~21.1]{pgm-koller-2009}---and, the uniform redraw
having full support \citep[cf.\ Prop.~6.13]{eci-peters-2017}, the particle
average is, at the level of the compared distributions, an exact test of
direct causal effect: zero when $p$ is not a direct parent of $q$, positive
when it is and the randomization reaches a witnessing context. It is thus
a direct-edge estimand, carrying no mediated influence; what survives of
its readings at lag $\geq 2$ is measured in Section~\ref{sec:lag}.
Thresholding the per-position CMI matrix at $\tau$ and projecting onto
event types yields the \emph{sample summary causal graph} of $s$ (paper's
Definition~3.2): a type edge is called if any supporting position pair
crosses $\tau$.

\paragraph{Ground truth and metric.} The synthetic generator (paper's
Eq.~21, specified in its evaluation appendix)
produces sequences from a known lagged mechanism over $\Xcal$ with lag
horizon $h$. Interventional truth is defined \emph{per sequence}: each
position is replaced by uniform draws over real event types (10
counterfactuals), the generator's conditionals are recomputed at positions
$q \in (p, p+h]$, and a type edge is true iff the mean KL divergence
exceeds $\delta = 0.05$. This truth is itself a benchmark construction:
it evaluates the direct effect at the factual mediator
configuration---exactly computable from the generator's conditionals and
inexpensive at benchmark scale---whereas a factual configuration can
mask a direct edge that a witnessing context elsewhere would reveal,
which is what the estimator's full-support randomization guards against
in deployment, where no truth is available. The two constructions thus
answer the same direct-effect question at different mediator
configurations; Section~\ref{sec:lag} measures the consequences.
Paper's Table~2 reports the mean per-observation $F1$ of
the sample summary graph against this truth. We stress, because it cost us
a debug cycle, that the metric is per-observation---one sequence per
summary graph, no corpus-level aggregation. A corpus-level graph is a
genuinely different and harder object, outside this note's scope.

\paragraph{Threshold guidance in the paper.} Four values bear on $\tau$ at
$|\Xcal| = 1000$: the paper's Table~2 constant ($3\times10^{-5}$); the
sensitivity-sweep optimum ($\approx 1.4\times10^{-5}$, paper's Fig.~9); the
scaling law $\tau_{\mathrm{opt}} \approx C\,|\Xcal|^{-0.96}$ with
$C = 1.72\times10^{-2}$, fit across $|\Xcal| \in \{100, \dots, 2000\}$ at
generator noise $\varepsilon = 0.04$ (paper's Fig.~10), giving
$1.7$--$2.3\times10^{-5}$; and the $10^{-4}$ of the qualitative
experiments (paper's Figs.~4--5). The paper thus treats $\tau$ as a swept,
setup-dependent quantity, and our results below are best read as an
elaboration of that stance.

\section{Reimplementation and setup}
\label{sec:setup}

\paragraph{Reimplementation.} We reimplemented the full pipeline from
scratch---generator, tokenizer, pretraining, staircase CMI,
thresholding, graph projection, evaluation---transcribing the staircase
and Bernoulli-KL algebra from the reference implementation
\citep{seq2cause}.

\paragraph{Worlds.} Two synthetic regimes, both at $|\Xcal| = 1000$,
$L = 64$, a decoder-only backbone at the paper's scale, and $N = 128$
particles. The \textbf{redundancy-matched regime} (the replication target)
uses generator sparsity $0.9$, lag horizon $h = 6$, truth margin
$\delta = 0.05$ with 10 counterfactuals, and a mechanism weight scale tuned
so the measured redundancy matches the
paper's benchmark value $R \approx 0.58$ (measured $R = 0.593/0.590$,
seeds 0/1); context $c = 6$ primary, $c = 20$ secondary. We match the
paper on the redundancy axis only; other generator settings, such as the
noise level, differ, which is part of why thresholds should be expected to
move. The \textbf{high-redundancy regime} uses the same generator family
at a mechanism scale yielding $R = 0.93$, $h = 3$, $c = 20$, $\sim$300k
training sequences; it serves as an out-of-regime probe. Two follow-up
worlds at $|\Xcal| = 10^{4}$ and $2\cdot10^{4}$, run later under the
same generator family and protocol (R-matched, default capacity), are
quoted where they extend a finding. A last family, used only in
Sections~\ref{sec:lag} and~\ref{sec:benchmark}, generalizes the
generator's hard-coded lag decay
to $e^{-r(k-1)}$ at $|\Xcal| = 1000$ ($r \in \{0.3, 0.1, 0\}$; the
paper's Eq.~21 is $r = 1$), each world redundancy-re-matched under the
same protocol.

\paragraph{Threshold selection and evaluation.} Thresholds live on a
fixed logarithmic grid from $10^{-6}$ to $10^{-1}$ containing the printed
constant and the paper's scaling-law anchor (12 points; the vocabulary
and particle sweeps use a 23-point superset). One $\tau$ per seed is
chosen by maximizing mean per-sequence $F1$ on the \emph{validation}
split and only then applied to the test split; test-side grids appear
only as sensitivity curves. Evaluation is mean per-sequence $F1$
against exact per-sequence interventional truth, over 1{,}024 head test
sequences per seed for the replication and sweep worlds ($64\times$ the
paper's per-run evaluation width) and 5{,}000 per seed for the
high-redundancy world. Two seeds throughout; cross-seed gaps were
$\leq 0.03$ everywhere.

\section{Replication at the paper's regime}
\label{sec:headline}

\begin{table}[t]
\centering
\small
\begin{tabular}{@{}llccc@{}}
\toprule
Reading & $\tau$ & $c$ & $F1$ (seed 0) & $F1$ (seed 1) \\
\midrule
Val-selected (blind) & $3\times10^{-2}$ & 6 & \textbf{0.9008} & \textbf{0.9051} \\
Grid optimum (test) & $3\times10^{-2}$ & 20 & 0.9004 & 0.9061 \\
Paper's constant & $3\times10^{-5}$ & 6 & 0.4553 & 0.4647 \\
Paper's constant & $3\times10^{-5}$ & 20 & 0.4650 & 0.4742 \\
\midrule
\multicolumn{3}{@{}l}{Published reference \citep[Table~2]{math2026trace}} &
\multicolumn{2}{c}{$\approx 0.91$} \\
\bottomrule
\end{tabular}
\caption{Matched-redundancy replication ($|\Xcal|=1000$, $L=64$, $h=6$,
$R \approx 0.59$; exact per-sequence interventional truth; 1{,}024 test
sequences per seed). Blind validation selection picks
$\tau = 3\times10^{-2}$ on both seeds and at both context widths.}
\label{tab:armc}
\end{table}

The results are shown in Table~\ref{tab:armc}: with $\tau$ selected blindly on validation data the
pipeline reaches $F1 = 0.9008/0.9051$ against the original paper's Table~2 $\approx 0.91$---a 
clean replication on an independent reimplementation, against exact
interventional truth, at the paper's own redundancy, and stable in the
context width. The selected threshold, however, is $3\times10^{-2}$, three
orders of magnitude above the value used for the paper's own setup; at
that value our runs score $F1 \approx 0.46$, consistently across context
widths and seeds. The optimal operating point is therefore reproduced for
the \emph{estimator} but not for the \emph{constant}: blind validation
selection recovers the published performance, while the threshold value
attaining it does not transfer between setups---a regime dependence that
the paper's own sensitivity analysis anticipates, and that
Section~\ref{sec:scaling} quantifies.

\section{How the threshold moves across regimes}
\label{sec:scaling}

\paragraph{A vocabulary sweep.} To measure the dependence directly, we
built new generators and pretraining per $(|\Xcal|, \text{seed})$ for
$|\Xcal| \in \{100, 500, 2000\}$, identical to the replication world
except $|\Xcal|$ and the weight scale (re-matched to
$|R - 0.59| \leq 0.015$); $\tau^{*}$ is the validation argmax, applied
once to the test split, and the exponent of
$\tau^{*}(|\Xcal|) \propto |\Xcal|^{b}$ is a pooled least-squares fit. A
companion arm repeats the three new sizes at the paper's own fixed
weight scale ($R$ floating), holding the mechanism family fixed as
$|\Xcal|$ varies.

\begin{table}[t]
\centering
\small
\resizebox{\textwidth}{!}{%
\begin{tabular}{@{}rcccccc@{}}
\toprule
$|\Xcal|$ & $R$ (s0/s1/s2) & truth pairs/seq & $\tau^{*}$ (grid, s0/s1/s2) & $\tilde\tau^{*}$ (median) & test $F1$@$\tau^{*}$ (s0/s1/s2) & $F1$@$C/|\Xcal|$ \\
\midrule
100  & .594/.592/.582 & 30--39 & 0.10 / 0.178 / 0.10 & 0.14 & .883/.859/.859 & .245/.254/.249 \\
500  & .591/.592/.595 & 49--51 & 0.0562 $\times 3$ & 0.046 & .908/.889/.893 & .301/.305/.296 \\
1000 & .593/.590/.603 & 50--52 & 0.03 / 0.0562 / 0.03 & 0.034 & .901/.905/.897 & .375/.384/.393 \\
2000 & .601/.598/.590 & 50--51 & 0.03 $\times 3$ & 0.029 & .899/.896/.900 & .474/.487/.482 \\
\midrule
\multicolumn{7}{@{}l}{pooled exponent over the swept range (12 points, three seeds): $b = -0.552$ (per seed $-0.539/-0.557/-0.559$)} \\
\multicolumn{7}{@{}l}{companion arm (fixed weight scale, $R$ floating $0.42$--$0.60$): $b = -0.489$ ($-0.495/-0.484/-0.488$)} \\
\bottomrule
\end{tabular}}
\caption{Vocabulary sweep, redundancy-matched arm ($h = 6$,
$c = 6$, $N = 128$; 1{,}024 sequences per split; exact per-sequence
truth). $\tau^{*}$ is the validation argmax;
the test-side argmax coincided with it on all 21 test replicas of both
arms. ``Truth pairs/seq'' counts type pairs after the Def.-3.2
projection; the position-level exact truth is flat at 50--53 pairs per
sequence at every $|\Xcal|$ (the rise from 30 reflects type collisions
at $|\Xcal| = 100$).}
\label{tab:m7}
\end{table}

\paragraph{Outcome.} As shown in Table~\ref{tab:m7}, the selection protocol replicates
at every size---test $F1$ at the selected $\tau^{*}$ is $0.86$--$0.91$,
with validation-to-test drift within $\pm 0.003$---while any single
fixed constant does not: transplanting a constant fit in one regime reads
$0.25$--$0.48$ here. Within the swept range the optimum falls with
vocabulary size much more slowly than $1/|\Xcal|$ (pooled $b = -0.552$),
and the log--log curve is concave (local slope $-0.88$ at
$|\Xcal| = 100$ easing to $-0.19$ at 2000).

\paragraph{What sets the scale: the $\delta$-straddle.} Per-pair
analysis of the emitted traces (32 per world, the generator's exact
per-pair KL replayed beside each estimate) shows that the optimum is
pinned to the truth margin. At every vocabulary size the true pairs
$\tau^{*}$ gives up have generator KL just above the margin
$\delta = 0.05$ (median $0.06$--$0.08$) and the false pairs it accepts
have KL just below it (median $0.02$--$0.03$---real but sub-margin
dependencies); from $|\Xcal| = 500$ up all of both are lag-1 pairs, and
the estimator reads lag-1 pairs at their KL (median
$\log_{10}(\mathrm{CMI}/\mathrm{KL})$ within $\approx 0.1$ in every KL
band). Both error tails are therefore pinned to $\delta$, and what moves
$\tau^{*}$ along the sweep changes: from $|\Xcal| = 100$ to $500$ it is
the collapse of the noise floor on null pairs ($\propto |\Xcal|^{-1.2}$
in median); from $500$ to $2000$ it is the true-edge bulk drifting
toward $\delta$ with the token-probability scale. Consistently,
$\tilde\tau^{*}$ decelerates (segment slopes $-0.70 / -0.41 / -0.28$,
curvature $+0.27 \pm 0.01$ over three seeds) toward a floor of order
$\delta/2$ rather than following any power law. The companion arm, which
holds the mechanism family fixed, differs only on the first segment
($b = -0.489$ vs $-0.552$): mechanism-scale conditioning is real but
small.

\paragraph{An out-of-sample test.} The two readings predict differently
at the next size, so we wrote the predictions down before the run:
$\tilde\tau^{*}(5000) > 0.021$ if the optimum is margin-pinned, $0.015$
from our $|\Xcal|^{-0.552}$ fit. The run read $\tilde\tau^{*} = 0.0230$, with the
straddle signature unchanged (missed-truth KL median $0.086$,
accepted-FP KL $0.027$, false positives $100\%$ lag-1): the power law
fit to our own twelve points is refuted one size out of sample, and the
margin reading holds. The residual movement is itself measurable---the
estimator's level on margin-band pairs, $\approx 0$ at
$|\Xcal| = 1000$--$2000$, drifts to $-0.16$/$-0.20$ decades at 5000---so 
even the margin-pinned threshold inherits a slow calibration drift,
to be re-measured per world rather than extrapolated.

\paragraph{Reading.} The paper's $\tau = C/|\Xcal|$ law is a faithful
fit within its own regime; what generalizes across regimes is the margin
anchor. Writing $b_1$ for the estimator's measured level on margin-band
pairs, the blind optimum sits within $\times 1.7$ of
$(\delta/2) \cdot 10^{b_1}$ at every size from 500 up, and within
$\times 1.5$ from 1000 up (per-seed $\times 1.5$--$1.65$ at
$|\Xcal| = 500$, where the noise floor is still receding; at
$|\Xcal| = 100$ the still-collapsing noise floor, not the margin,
sets the optimum); the follow-up worlds read $\times 1.07$ at $10^{4}$
and $\times 0.70$ at $2\cdot10^{4}$, where no absolute
$\tau(|\Xcal|)$ fit survives. Across 27 world-runs spanning six
vocabulary sizes and three same-size capacity regimes, $b_1$ is
predicted by the model's residual fit error (the excess of final
validation loss over the generator's conditional-entropy floor;
$R^{2} = 0.78$) better than by $\log|\Xcal|$ ($R^{2} = 0.67$): the
calibration drifts as the modeling problem gets harder, whatever the
vocabulary. The operational conclusion, the authors' as well as ours, is
that the transferable object is the selection protocol---``$\tau$
selected on held-out validation data by maximizing $F1$'' reproduces the
paper's number at every size we tried---and that a quoted threshold is
best stated relative to the truth margin and the estimator's measured
calibration, not as an absolute constant.

\section{What TRACE recovers: a lag-graded recall limit of the staircase
construction at a single global threshold}
\label{sec:lag}

\paragraph{Recall is the lag-1 share.} Every world in this section up
to the flat-decay control of the closing paragraph carries the synthetic
generator's default lag decay ($e^{-(k-1)}$, the $r = 1$
member of the Section~\ref{sec:setup} family)---the regime the paper's
own evaluation runs in, and the default whose skew
Section~\ref{sec:benchmark} measures. At the selected single global
threshold the estimator recalls lag-1 true edges (cause immediately
before effect) at
$0.97$--$0.99$ on every world in the sweep, while true edges at lag
$\geq 2$ are recalled at $0.15$ ($|\Xcal| = 100$), then $0.006$, $0.004$
and $0.000$; false positives are $98$--$100\%$ lag-1 from
$|\Xcal| = 500$. Overall recall therefore equals the lag-1 share of the
truth on every world ($0.81$--$0.84$ observed vs.\ lag-1 shares
$0.80$--$0.85$). The mechanism is a level gap, not sampling noise: lag-1
pairs are read at their generator KL, while lag $\geq 2$ true pairs read
from one to over three decades below it as $|\Xcal|$ grows, three
decades at $R = 0.93$ (median $\log_{10}(\mathrm{CMI}/\mathrm{KL})$
$-3.0$ at lag 2, $-3.3$ at lag 3), and about five decades in the
follow-up worlds at $1$--$2\cdot10^{4}$, where lag $\geq 2$ recall
is exactly $0.000$. No \emph{single global} threshold can rescue a
class that reads decades below the lag-1 scale the threshold is fit
on. The class is, however, not wholly invisible: selecting $\tau$ per
lag (validation argmax restricted to lag-$k$ candidates) recovers
$0.35/0.47$ of lag-2 truth on the same replication runs, at
$\tau_2 \approx 4$--$8\times10^{-4}$---two orders of magnitude below
the pooled optimum---while lags $\geq 3$ remain unrecovered by any
threshold family selectable on these runs: their truth reads below the
lag's own null floor, and the validation cells (6--23 true pairs) are
too thin to select a threshold from.

\paragraph{Explaining the high-redundancy ceiling.} In the
high-redundancy world ($R = 0.93$) blind val-selected $\tau$ reaches
per-sequence $F1 = 0.8271/0.8277$ against exact truth (5{,}000 test
sequences per seed; cross-seed gap
$0.0007$), against $0.90$--$0.91$ at $R \approx 0.59$. The lag
decomposition accounts for the ceiling: lag-1 truth is saturated (recall
$1.000$; $\approx 99\%$ of lag-1 candidate pairs are true), the missed
truth is entirely lag-2/3 pairs with margin-band KL (median $0.11$), and
the selected $\tau^{*} = 10^{-2}/3\cdot10^{-3}$ is the same
$\delta$-straddle one lag class down and three decades 
lower---$61$--$70\%$ of its false positives are lag-2 pairs. What the redundancy
changed is the truth's lag composition ($0.70/0.21/0.09$ at lags
$1/2/3$, vs.\ $0.84$ lag-1 in the replication world): per-sequence
recall is again the recallable-lag share. The practical point is one
of score interpretation: the same protocol, reading each lag class
the same way in both worlds, scores $0.91$ where truth is $84\%$
lag-1 and $0.83$ where it is $70\%$ lag-1, so the aggregate $F1$
carries the domain's lag composition as much as the estimator's
quality, and a headline-score difference across datasets is not by
itself evidence of better or worse recovery. Only per-lag recall
separates a shift in the truth's composition from a change in the
estimator.

\paragraph{The mechanism: mediator randomization off the data
distribution.} The synthetic evaluation
uses two different methods for the ground-truth discovery and for one-shot 
estimation at inference time. The ground-truth intervention is atomic 
and factual: only the cause position is re-drawn, and every other position, 
including the mediating events, is held at its observed value. Conditioning on the
observed mediators blocks the mediated pathway, so a lag-$\geq 2$ truth
edge is a \emph{direct} dependence, evaluated with the effect's real
context in place. The estimator's randomized do-operator (Def.~4.6)
re-draws the entire window tail after the intervention point, so the
same observed effect token is predicted, on \emph{both} sides of the CMI
contrast, under a context of uniform noise---a mediator configuration
far off the distribution the sequences are drawn from. Stripped of the real
predecessors it was generated from, that token's factual probability
collapses toward its marginal rarity, and the CMI read from it falls by
decades, whatever the strength of the direct dependence and at any
particle count. (In intervention-semantics terms both constructions ask
the direct-effect question---the truth evaluates it at the factual
mediator configuration, the estimator averages it over uniform mediator
draws---so the deep-lag deficit is a loss of reading scale, not a
change of estimand. The factual configuration can itself mask an
edge the randomization would witness: where the observed mediator
values happen to suppress a direct dependence, the truth records no
edge, and a real direct edge the one-shot estimate recovers is
scored a false positive---the error is then the truth's, not the
estimator's.) The gap carries the fingerprints of this
probability-scale mechanism rather than of an estimation error or a
scope exclusion: it widens with vocabulary (from under one decade at
$|\Xcal| = 100$ to five--six decades at $1$--$2\cdot10^{4}$, where
the observed token is rarer under noise), and it moves by decades at
fixed $|\Xcal|$ when the marginal distribution or the model's achieved
fit changes. Consistently, an atomic variant of the estimator---noising
only the candidate cause row, mediators left at their observed values,
i.e.\ the truth's own construction run as an estimator---recovers
$0.45/0.55$ of the lag-2 truth the staircase misses at a
single pooled $\tau$ (equal-particle paired rescore), because the
factual context keeps the effect token's probability at its observed
scale; lags $\geq 4$ stay at zero under either construction. We use it
as an instrument sizing the staircase's share of the deficit, not as
a rival estimator. The operational implication for users of the
algorithm in its published form, on data whose causal strength decays
with lag: at a single global threshold, the graph TRACE recovers is a
direct, adjacent-influence
(lag-1) graph. Longer-range dependencies then appear only insofar as they
are chains of lag-1 links, and per-lag recall should be reported so
that the unrecovered class is visible.

\paragraph{A flat-decay control isolates the staircase's own
share.} To size that share independently of the data's lag profile,
we generalize the generator's decay to $e^{-r(k-1)}$ and re-match
redundancy (Section~\ref{sec:setup}). On a flat world ($r = 0$; truth
now $15\%$ lag-1 and $\approx 1.9\times$ denser), the same pipeline at
its single validation-selected $\tilde\tau^{*} = 0.041$ recalls truth
at $0.87/0.76/0.69/0.63/0.59/0.40$ across lags 1--6 (test side,
seed-replicated within $0.03$ per lag): the limit is graded, not
absolute. The $\delta$-straddle of Section~\ref{sec:scaling} transfers
to the flat world ($\tilde\tau^{*}$ again between $\delta/2$ and
$\delta$; missed-truth KL median $0.15 \in [\delta, 3\delta]$), though
its false positives are no longer lag-1-dominated ($34\%$). (The flat
worlds run hotter on the residual fit error of
Section~\ref{sec:scaling}, $\hat\varepsilon \approx 0.19$ vs.\ $0.13$
in the sweep, consistent with their denser truth; we disclose this
since calibration tracks $\hat\varepsilon$.) Even with lag-uniform
truth, staircase recall declines with lag ($0.87 \to 0.40$ above),
while the atomic variant run on the same emits is essentially
lag-flat---the staircase reads $13$ pp \emph{above} atomic at lag 1,
parity at lag 2, and $8$--$26$ pp below it at lags 3--6 (each
construction at its own validation-swept pooled threshold). This
lag-graded residual is the staircase's own share of the deficit, and
the only part of it that belongs to the algorithm. A second share
belongs to the single shared threshold: it is the part the per-lag
$\tau$ family above recovers. The third and the dominant share---the data's own
decay---belongs to the benchmark, and is estimated in the next section.

\section{What validating on the default benchmark certifies: a practitioner warning}
\label{sec:benchmark}

\paragraph{A doubly skewed default.} The synthetic benchmark is
specified in the paper's evaluation appendix (Eq.~21) and is not a part
of the algorithm per se, but it is where every headline number comes from.
Its mechanism weights hard-code the exponential lag decay
$e^{-(k-1)}$, and at that default the benchmark is skewed toward
lag 1 twice over: $\approx 85\%$ of position-level interventional
truth sits at lag 1, and the deep-lag truth that remains reads below
the estimator's noise floor---on our runs, below the lag's own null
floor from lag 3. The skew is dose-dependent, not incidental: recall
at every deeper lag falls monotonically as the decay rate rises
toward the default $r = 1$ (at lag 4:
$0.63 \to 0.43 \to 0.06 \to 0.00$ for $r = 0, 0.1, 0.3, 1$).

\paragraph{What validating on it certifies---and conflates.} Headline
per-sequence $F1$ on this benchmark is attainable by lag-1 recovery
alone: our replication scored $0.90$--$0.91$ while recalling
essentially no lag $\geq 2$ truth. Validation against the default
generator therefore certifies adjacent-influence recovery only---not
the long-range dependencies that motivate event-sequence causal
discovery, including the paper's own motivating examples (server
logs, medical records), where nothing guarantees that influence
concentrates at lag 1. And it conflates the two findings this note
separates: a practitioner who validates only on this default benchmark observes zero
lag-$\geq 2$ recall and may conclude, as we nearly did, that the
method cannot recover it at all---although the dominant share of that
deficit belongs to the benchmark's decay compounded with the shared
threshold, not to the staircase construction, whose own share is the moderate,
lag-graded residual of Section~\ref{sec:lag}. The flat-decay
generalization is a one-parameter control any user can run to
separate the two on their own deployment.

\section{The particle budget}
\label{sec:particles}

The paper reports diminishing returns in $N$ with no significant change
beyond $N = 256$ (paper's Fig.~8); the authors' expectation, when asked, is
saturation around $128$, since density-estimation error should dominate
the Monte-Carlo error of the particle average. We tested this
at the paper's regime: inference-only re-runs of the frozen replication
models (1{,}024 test sequences, $c = 6$) at $N \in \{2, 4, \dots, 512\}$,
read at the selected $\tau^{*} = 0.03$, with $N_{\mathrm{sat}}$ defined in
advance as the smallest
$N$ such that $F1$ and recall stay within $0.01$ of their $N = 512$ values
for every larger $N$, on both seeds. \textbf{Outcome:
$N_{\mathrm{sat}} = 2$}---a property of our operating point, not a
universal constant; the transferable object is the rule this section ends
with. Over the whole sweep $F1$ spans $0.0012$
($0.8999 \to 0.9011$ / $0.9042 \to 0.9054$) and recall $0.0015$, while
inference wall time is linear in $N$ (47\,s at $N = 2$ and 3{,}320\,s at
512).

Because flat-from-two invites the suspicion of a degenerate model, we
diagnosed it on the persisted per-pair CMI matrices. \emph{The estimator
is not saturated; the observable is.} On exact-truth position pairs
$|\mathrm{CMI}_N - \mathrm{CMI}_{512}|$ falls like $N^{-0.46}$ on both
seeds---textbook Monte-Carlo behaviour---but the true-pair median sits
$1.6$ decades above $\tau^{*}$ and the non-true median $3$ decades below
it, so the noise almost never crosses the threshold: 13/12 of 28{,}320
position classifications change between $N = 2$ and $512$, against
$1{,}784/1{,}723$ at $\tau = 3\times10^{-5}$ and $12{,}211/12{,}459$ at
$10^{-6}$. \emph{The published reference points are one curve:} applying
the same rule at every grid $\tau$ gives $N_{\mathrm{sat}} > 256$ for
$\tau \in [1.7, 3.4]\times10^{-5}$, $128$--$256$ up to
$1.7\times10^{-4}$, $64$ at $3\times10^{-4}$, $8$ at $10^{-3}$, and $2$
for every $\tau \geq 1.8\times10^{-3}$---the paper's ``$\geq 256$'', the authors'
``$\approx 128$'' and our ``2'' are indexed by the threshold's distance
to the non-edge noise floor. (Inside the floor, the right-skewed
non-negative per-particle KL means \emph{more} particles give \emph{more}
false positives: precision at $3\times10^{-5}$ falls from $0.45$ at
$N = 2$ to $0.27$ at 512.) \emph{It is not model collapse:} the model's
predictive entropy is above the generator's exact conditional entropy
(median $3.10/3.14$ vs.\ $2.88/2.83$ nats), and $\approx 94\%$ of the
particle variance sits on the intervention side. The authors' reading is
confirmed with one refinement: against the generator's exact per-pair KL
the estimate has an $N$-independent error floor of $0.7$--$0.9$ nats RMS
while the Monte-Carlo part is $0.40$ nats at $N = 2$ and $0.05$ at 
256---density-estimation error dominates, and at a threshold three decades
above the floor it does so from $N$ of a few. The general rule:
\emph{set the particle budget by the selected threshold's distance to the
noise floor, not by a fixed $N$.}

\section{Guidance for practitioners}
\label{sec:practitioners}

For anyone applying TRACE to their own event sequences---or rebuilding
it and validating the rebuild---our findings reduce to five rules.

\begin{enumerate}
\item \textbf{Select $\tau$ on held-out data, and expect it at the
  truth margin.} The validation sweep is inference-only---no
  retraining---and in our hands it is what carries the paper's
  performance to every regime we tried, while any fixed constant
  transplanted across regimes loses $0.4$--$0.6$ $F1$, and even a power
  law fit to our own sweep failed one size out of sample. The number the
  sweep returns has a meaning: for a level-calibrated model the optimum
  sits at roughly $\delta/2$ to $\delta$---half to one of the
  effect-size margin that defines an edge worth calling---and its
  errors concentrate on both sides of that margin. Quote thresholds
  relative to the margin and the measured calibration, never as
  absolute constants, and expect the calibration to drift as the
  modeling problem gets harder.
\item \textbf{Size the particle budget from the selected threshold's
  margin, not from a fixed $N$.} Once $\tau^{*}$ is known, a handful of
  particles suffices if it sits decades above the non-edge noise floor---in 
  our runs $N = 2$ matched $N = 512$ at $70\times$ less compute.
  Large particle counts pay only when the threshold approaches the
  floor, and \emph{inside} the floor extra particles actively increase
  false positives.
\item \textbf{Know which observable you are computing.} TRACE's headline
  metric is per-observation: one summary graph per sequence, scored per
  sequence. Aggregating evidence across a corpus into a single
  system-level graph is a different and harder estimation problem, and
  per-sequence numbers do not transfer to it.
\item \textbf{Under lag-decaying data and a single global threshold,
  read the graph as direct, adjacent (lag-1) influence---and report
  per-lag recall.} Lag-1 true
  edges are recalled at $0.97$--$0.99$; true edges at lag $\geq 2$ read
  decades low and are recalled near zero at any single global
  threshold, because the tail-randomized context collapses the observed
  effect token's factual probability. Overall recall is then the lag-1
  share of the domain's true structure---our $\approx 0.83$ ceiling at
  extreme redundancy is exactly that share falling---and aggregate
  $F1$ cannot distinguish that fall from estimator degradation.
  Before concluding the
  deeper signal is absent, sweep $\tau$ per lag: it recovered a third
  to a half of lag-2 truth in our runs, and the selected $\tau_k$
  ladder is itself diagnostic (it falls ${\sim}500\times$ over lags
  1--6 under strong decay, ${\sim}30\times$ when flat). If long-range
  recovery is the claim, run a flat-decay variant of the generator as
  a control---on ours, one validated threshold recalled every lag.
  Reporting per-lag recall makes the invisible class visible.
\item \textbf{Validate long-range recovery on a slowed or flattened
decay, not on the default generator alone.}
The default generator's hard-coded lag decay ($e^{-(k-1)}$ in the
  mechanism weights) concentrates $\approx 85\%$ of interventional
  truth at lag 1 and pushes deep-lag truth below the estimator's own
  noise floor, so the headline per-sequence $F1$ is attainable by
  lag-1 recovery alone: our replication scored $0.90$--$0.91$ while
  recalling essentially no lag $\geq 2$ truth. A rebuild validated
  only against this generator is therefore certified for
  adjacent-influence recovery, not for the long-range dependencies
  that motivate event-sequence causal discovery. To test the latter,
  slow or flatten the decay (one generator parameter,
  Section~\ref{sec:benchmark}) so every lag carries recallable truth, and
  read per-lag recall against both the staircase and the atomic
  construction: on our flat control the staircase's residual,
  lag-graded deficit ($8$--$26$ pp below atomic at lags 3--6) is the
  algorithm's own share, cleanly separated from the generator's.
\end{enumerate}

\section{Conclusion}

Scored its own way and given its own regime, TRACE does what its paper
says: an independent reimplementation reaches $F1 = 0.90$--$0.91$ against
exact interventional truth, matching paper's Table~2, and $0.86$--$0.91$ at every
vocabulary size from 100 to 2000 under the validation-sweep protocol the
paper's own sensitivity analysis licenses. Behind those numbers sit
three regime facts and one warning. The selected threshold converges
on the truth
margin: its errors straddle $\delta$, its residual movement is the
estimator's slowly drifting calibration, and a power-law extrapolation
of it fails one size out of sample. At a single global threshold on
lag-decaying data, the recovered graph is a direct, adjacent-influence
graph: no such threshold reaches the class that reads decades below
the lag-1 scale, a per-lag family recovers the nearest of it, and the
staircase's own share of the deficit---isolated on a flat-decay
control the pipeline otherwise recovers at every lag---is a moderate,
lag-graded residual. The
particle budget is set
by the selected threshold's margin over the noise floor, not by a fixed
count. The warning: the paper's default benchmark is specified in its
appendix and skewed toward lag 1 in both truth mass and CMI scale;
validating on it certifies lag-1 recovery only and conflates the
benchmark's skew with the staircase's limit---validate long-range
claims on a slowed or flattened decay. \emph{Thresholds, budgets and recall ceilings are regime
properties; the protocol---select on validation, quote against the
margin, report per-lag recall, and control the generator's decay when
long-range claims are at stake---is the method property practitioners
should carry.}

\subsection*{Reproducibility statement}

All experiment plans were committed before the runs they govern; gated
numbers were produced under validation-frozen thresholds recorded in a
versioned registry prior to any test-side run; outcome bands, saturation
rules and escalation rules for the vocabulary and particle sweeps were
committed before any of their runs existed. Experiments ran on single
NVIDIA A10G instances (pretraining $\sim$30--40 min per world; the
particle sweep inference-only); the runs behind this note sum to under
40 GPU-hours excluding boot and setup, and the pipeline is
bit-reproducible per configuration and seed on this hardware. 

\subsection*{Acknowledgments}

We contacted Hugo Math and Rainer Lienhart on 2026-08-14 describing the
replication findings and asking about the threshold scale. Hugo Math
replied on 2026-08-17 with clarifications that shaped
Sections~\ref{sec:scaling}--\ref{sec:particles} (implementation checks
we then ran, the local-sharpness
hypothesis, and the particle-count expectation); we are grateful for the care of that engagement
and cite the reply as personal communication with the authors' permission. The per-pair tail
analyses behind Sections~\ref{sec:scaling} and~\ref{sec:lag}---which
deepen the authors' local-sharpness hypothesis into the margin-straddle
reading---and the particle-count diagnosis were run in response to
that exchange. Two further replies (2026-08-24 and 2026-08-26) shared
per-lag-threshold and flat-vs-decayed control experiments on the
authors' own generator; the per-lag threshold family and the
flat-decay control of Sections~\ref{sec:lag}
and~\ref{sec:benchmark} replicate those designs on
ours.

\bibliographystyle{plainnat}
\bibliography{references}

\end{document}